\pdfoutput=1

\documentclass[11pt]{article}

\usepackage{acl}

\usepackage{times}
\usepackage{latexsym}
\newcommand{\RNum}[1]{\uppercase\expandafter{\romannumeral #1\relax}}
\newcommand{\zh}[1]{\begin{CJK}{UTF8}{gbsn}#1\end{CJK}}
\newcommand{\zhkai}[1]{\begin{CJK}{UTF8}{gkai}#1\end{CJK}}
\usepackage[linesnumbered,ruled]{algorithm2e}
\usepackage{amsmath}

\usepackage[T1]{fontenc}

\usepackage[utf8]{inputenc}

\usepackage{microtype}
\usepackage{CJK}
\usepackage{graphicx}
\usepackage{hhline}
\usepackage{booktabs}

\title{Nominal Metaphor Generation with Multitask Learning}

\author{Yucheng Li~\textsuperscript{1}, Chenghua Lin~\textsuperscript{2}\thanks{~~Corresponding author}~, Frank Guerin~\textsuperscript{1} \\
\textsuperscript{1}~Department of Computer Science, University of Surrey, UK \\
\texttt{\{yucheng.li,f.guerin\}@surrey.ac.uk}\\ \textsuperscript{2}~Department of Computer Science, University of Sheffield, UK \\
\texttt{c.lin@sheffild.ac.uk }
}

\begin{document}
\maketitle
\begin{abstract}
Metaphor generation is a challenging task which can impact many downstream tasks such as improving user satisfaction with dialogue systems and story generation. This paper tackles the problem of Chinese nominal metaphor generation by introducing a multitask metaphor generation framework with self-training and metaphor identification mechanisms. Self-training addresses the 
data scarcity issue of metaphor datasets. That is, instead of solely relying on labelled metaphor datasets which are usually small in size, self-training helps identify potential metaphors from a large-scale unlabelled corpus for metaphor generation.
The metaphor weighting mechanism enables our model to focus on the metaphor-related parts of the input (e.g., the comparison of the metaphor and comparator) during model learning and thus improves the metaphoricity of the generated metaphors. 
Our model is trained on an annotated corpus consisting of 6.3k sentences that contain diverse metaphorical expressions. 
Experimental results show that our model is able to generate metaphors with better readability and creativity compared to the baseline models, even in the situation 
where  training data is insufficient.
\end{abstract}

\section{Introduction}
Metaphor is commonly used in human language as an effective communication device. Typically, metaphors compare a concept or an object to another with the intent to make the expression more vivid, or to make unfamiliar things easier to understand \cite{paul1970-figurative}.

\begin{table}[t]
    \centering
    \begin{tabular}{|p{4.5cm}r|}
    \hline
         1. \zh{百合花\zhkai{好似}一瓶香水}. & \\
         Lilies smalls \textbf{like} a &\\ bottle of perfume. &  \textit{Nominal} \\ \hline
         2. \zh{他很擅长\zhkai{编织}人际关系}. & \\
         He is good at \textbf{weaving} &\\ relationship. & \textit{Verb} \\\hline
         3. \zh{银行股\zhkai{跳水}} & \\
         Bank stock \textbf{dive}. & \textit{Personification} \\ \hline 
         4. \zh{他可以像大厨一样烹饪} & \\
         He can cook like a pro. & \textit{Non} \\ \hline
    \end{tabular}
    \caption{Examples of different types of Chinese metaphor: \textit{nominal} metaphor, \textit{verb} metaphor, and \textit{personification} metaphors. Metaphorical words are bold.
    }
    \label{tab:CM_examples}
\end{table}

According to linguistic studies of Chinese, metaphors are particularly important in Chinese as metaphor is the dominant figurative language in Chinese \cite{wang2004}.
As shown in Table \ref{tab:CM_examples}, there are different types of Chinese metaphors. \textit{Nominal} Metaphors (NMs), also known as \zh{比喻} in Chinese, are figures of speech associating a noun with another noun through a comparator such as  and \zh{像,是,变成} (equivalent to \textit{like, be, become} in English). 
\citet{wang2004} claims that the nominal metaphor requires the comparison to be drawn from objects different in nature. Therefore, even though the fourth example in the table uses a classic comparator ``like'', it does not make it a metaphor as it compares a person to another person. \textit{Verb} metaphors are metaphors whose verbs are used metaphorically. The verb metaphor shown in Table \ref{tab:CM_examples} uses \textit{weaving}, a verb which is usually related to cloth or loom, to describe human relationships. The third type of Chinese metaphor is \textit{personification} (also known as \zh{拟人} in Chinese), which treats objects as human and can act like humans. 

Previous efforts on metaphor generation demonstrate the task can bring benefits to a wide range of NLG downstream tasks. \citet{glucksberg1989-metaphors-in-conversation} suggested that \textit{verb} metaphors are important to an engaging conversation. \citet{zhou2020-love} showed machine-generated NMs are effective in stimulating user interest in communicating with chatbots. \citet{chakrabarty2020-simile-gen,chakrabarty2021-mermaid} conducted human evaluations comparing literal expressions from machine-generated stories and poems with machine-generated metaphors and found users prefer the text with metaphors.

In this paper, we mainly focus on generating the \textit{nominal} metaphors. The generation of the NMs is defined as follows:
given the subject of the metaphor, i.e., ``\textit{Lilies}'' in the first example of Table \ref{tab:CM_examples}, generate a comparison containing the comparator and the object of the comparison, i.e., ``\textit{like}'' and ``\textit{a bottle of perfume}'' in the example, respectively.
There are two main challenges to the Chinese metaphor generation task. The first issue is the t annotated corpus. Existing Chinese metaphor corpora are not large enough to power current data-driven text generation approaches. Second, the auto-regressive fashion language modelling is ineffective for learning metaphor generation. Because a metaphor can be hidden in a very long sentence, the generative model tends to learn the entire sentence sequence rather than focusing on the metaphorical part of the input.

To address the aforementioned challenges, we propose a novel neural metaphor generation model that requires only limited labelled metaphor data for model training. This is achieved by a multitask framework which jointly performs novel self-training and metaphor weighting mechanisms. First, to tackle the scarcity issue of metaphor datasets, we employ self-training to leverage additional unlabelled data to improve the metaphor generation performance. Self-training consists of three main steps: (1) train a teacher model on labelled training data; (2) detect potential metaphors in the unlabelled corpus; and (3) train a student model on the combination of the labelled as well as newly identified  metaphors from the unlabelled data. 
Second, we propose to employ metaphor identification to reveal metaphor-related parts of the input. This permits our model to focus on the metaphor-related parts of the entire input sentence via assigning higher weights to metaphor-related content. Introducing metaphor identification not only improves the efficiency of the model training process, but also improves the metaphoricity of the generated metaphors. 

As there are limited data available for nominal metaphor generation, we collect and annotate two corpora for our model training, namely, Chinese Metaphor Corpus (CMC) and Chinese Literature Corpus (CLC). CMC contains 2.7k metaphor examples and 3.5k literal examples, which can be used for both metaphor detection and generation. CLC is a large-scale unlabelled Chinese literature corpus, which can be leveraged by our self-training algorithm for identifying additional high quality labelled metaphors. We conduct both automatic and human evaluation to evaluate our model's performance in metaphor generation. 
Experimental results show that our model is able to generate metaphors with better readability and creativity compared to the baseline models, even in the situation where training data is
insufficient. 
Source code and data can be found in \url{https://github.com/liyucheng09/Metaphor_Generator}. 

\section{Related Work}

Prior works on computational processing of metaphors can generally be classified into detection, interpretation and generation tasks. 

\subsection{Detection and Interpretation of Metaphors}

\citet{krishnakumaran2007-type-of-metaphor-and-NMs-detect} exploit the absence of a hyponymy relation between subject and object to identify metaphorical utterances. \citet{shlomo2015-NMs-detect-Random-Forest} propose a random forest-based classifier for NM identification using both conceptual features such as abstractness and semantic relatedness such as domain corpus frequency. \citet{su2016-meta-interpretation} follow the idea of hyponymy relationship absence from \cite{krishnakumaran2007-type-of-metaphor-and-NMs-detect} and implement it using cosine distance between pre-trained word2vec embeddings of the source and target concepts. \citet{liu2018-simile-detect,zeng2020-simile-detect} tackle Chinese simile detection by designing a multi-task framework and a local attention mechanism. Here, simile is a type of NM, which uses direct comparator such as ``like'' and ``as''. \citet{su2016-meta-interpretation,su2017-meta-interpretation} focus on \textit{Nominal} and \textit{verb} metaphor interpretation and perform experiments on English and Chinese metaphors. They extract properties of both the subject and object of the metaphors from WordNet and use pre-trained word2vec embeddings to identify related properties shared by the compared objects/concepts pair.

\begin{figure*}[ht]
    \centering
    \includegraphics[width=0.8\textwidth]{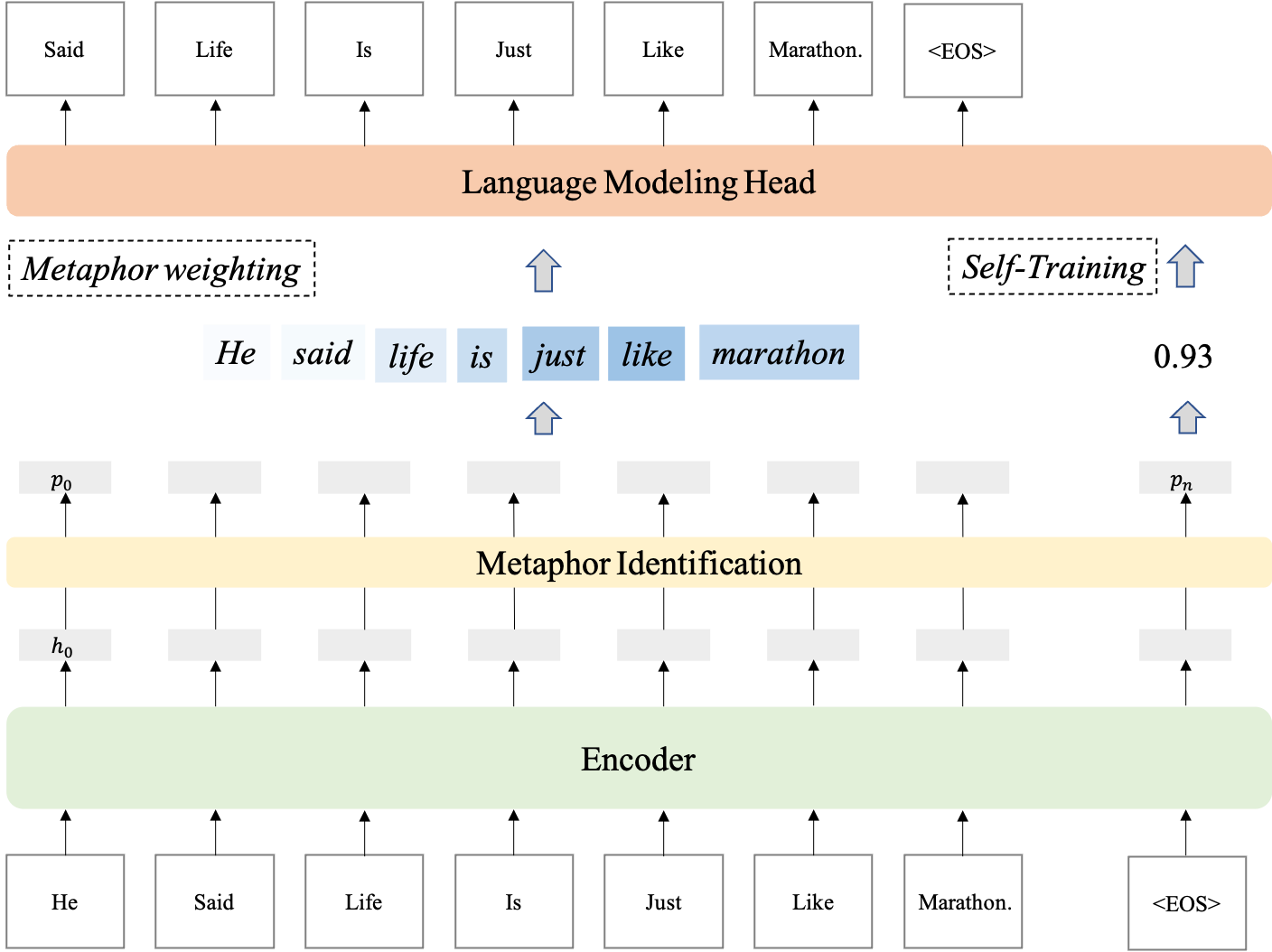}
    \caption{The overall framework of our method. The metaphor identification module performs self-training and metaphorical word identification.}
    \label{fig:framework}
\end{figure*}

\subsection{Generation of Metaphors}
Despite the benefits that metaphor generation can bring to many NLG tasks, works on this task are still relatively sparse. Early works for metaphor generation often rely on templates. \citet{terai2010-meta-gen-AIsLikeB} compute the relatedness between concepts with computational language analysis and select candidates to fill metaphor templates, e.g., ``A is like B''. \citet{veale2016-meta-gen-Gotham-City} uses a knowledge-base to generate $XYZ$ style metaphors such as ``Bruce Wayne is the Donald Trump of Gotham City''. \citet{zhou2020-love} not only chooses candidate concept pairs by word embedding similarity to fill the template but also chooses appropriate comparators to link the concept pair. \citet{chakrabarty2020-simile-gen} introduce a neural style transfer approach for simile generation, which fine-tunes a pre-trained sequence-to-sequence model on a literal-simile parallel dataset. Nevertheless, previous template-based approaches heavily constrain the diversity of generated metaphors, and both template methods and neural methods produce metaphors with a relatively simple structure.

\subsection{Metaphor Corpora}
Annotating metaphors can be quite challenging as cognitive scientists and psychologists define a variety of theories for metaphor. One popular approach to annotate metaphors in open text is the Metaphor Identification Procedure (MIP) introduced by \citet{steen2010-from-mip-to-mipvu-most-frequent-verbal}. Their study also releases the first metaphor dataset named VUAMC, which focuses on \textit{verb} metaphors and contains around 7k metaphor examples. VUAMC is widely used in metaphor identification and served as a benchmark in the first and second ACL workshop on figurative languages \cite{leong2018report,leong2020report}. \citet{chakrabarty2020-simile-gen} crawl web content and construct the first English simile corpus to train their metaphor generator.
\citet{liu2018-simile-detect} release a small  Chinese metaphor corpus with 120 examples including both \textit{verb} and \textit{nominal} metaphors. \cite{zeng2020-simile-detect} publish a Chinese simile dataset focusing on a specific NM which using the comparator ``like''.

\section{Methodology}

In this section, we provide the technical detail of our proposed framework metaphor generation. The overall model architecture is shown in Figure \ref{fig:framework}, which includes: (1) the GPT2 model \cite{radford2019-gpt2} and text prediction layer; (2) the metaphor identification module, which identifies potential metaphors from the unlabelled dataset (i.e., via self-training) and emphasises metaphorical words of the input sequence (i.e., via metaphor weighting).

\subsection{Text Modeling and Metaphor Identification Module}

\paragraph{\textbf{The Pre-Trained Language Model}} We employ GPT2, a pre-trained unidirectional transformer language model, as our basic encoder. 
Given a sentence $S=(w_0, \cdots, w_n)$, the GPT2 model produces a list of contextualized token embedding $(h_0, \cdots, h_n)$, where $h_i$ is the representation of the $i$-th input token $w_i$. Since the GPT2 model is a \textit{unidirectional} language model, the produced contextualized word embedding captures only the information of preceding context. 
For example, $h_i$ captures information asserted before the $i$-th token, which is $(w_0, \cdots, w_i)$. 

\paragraph{\textbf{The Text Prediction Layer}} To predict the output token, we apply a linear layer followed by a softmax function to the contextualized word embedding.
\begin{equation}
P(w_{i+1}|w_1, \cdots, w_i)=\mathrm{softmax}(Wh_i+b)
\end{equation}
where $W$ and $b$ are trainable weight matrix and bias for text generation, respectively; $h_i$ is the contextualised word embedding of the $i$-th word.

\paragraph{\textbf{The Metaphor Identification Module}} 
The metaphor identification module is used to assign metaphorical probability to \textit{sentences} (used in self-training) or \textit{sub-sentences} (used in metaphor weighting).
Specifically, we apply a linear layer plus a softmax layer on the contextualized embedding to compute the metaphorical probability. 
Formally, after obtaining 
word representations $(h_0, h_1, \cdots, h_n)$ by GPT2, we compute the metaphor probability as follow:
\begin{equation}
p_i=\mathrm{softmax}(W_mh_i+b_m)
\end{equation}
where $W_m$ and $b_m$ are the trainable weight matrix and bias for metaphor identification, and $p_i$ is the metaphor probability of the sub-sentence $w_0, \cdots, w_i$ (or the whole sentence if $i=n$). Because $h_i$ captures the contextual information preceding $w_i$, $p_i$ indicates whether the sub-sentence $w_0, \cdots, w_i$ contains metaphorical expression.


\subsection{Self-Training} Self-training is an effective approach to make use of additional data to enhance deep learning model performance. It has shown significant progress in many deep learning tasks, such as machine translation \cite{he2019NMT-Self-Training}, speech recognition \cite{parthasarathi2019self-training-speech-recognition}, and image classification \cite{xie2020self-training-imagenet}. As shown in \cite{he2019NMT-Self-Training}, self-training provides significant performance gains in sequence generation when the supervised corpus is relatively small. To ensure generation fluency and diversity, we employ self-training to train our metaphor generator.

The classic self-training procedure is shown in Algorithm \ref{algo1}. Classic self-training starts from a teacher model trained with a supervised dataset $U$. The teacher model is then applied to unlabelled data to obtain the pseudo label. Finally, we train a student model on the syntactic dataset incorporating the pseudo label dataset and the supervised dataset $S\cup U$.

\begin{algorithm}[t]
Train a base model $f$ on $U=\{x_i, y_i\}$\;
\Repeat{convergence or maximum iterations are reached}{
Apply $f$ to the unlabeled instance $C$ \;
Select a subset $S\subset\{(x, f(x))|x\in C\}$ \;
Train a new model $f'$ on $S\cup U$\;
}
\caption{Classic Self-Training}
\label{algo1}
\end{algorithm}

Instead of training a \textit{independent} teacher model, we embed the teacher model in the overall model as an attention mechanism. As shown in Figure \ref{fig:framework}, each unlabelled instance is weighted by its metaphorical probability during metaphor modelling. We use the metaphor identification module to score each sentence with the probability of being metaphorical. Formally, given a unlabelled sentence $x=(w_0, \cdots, w_n)$, the metaphor identification module compute its metaphorical probability $p_n$.
\begin{equation}
p_n=\mathrm{softmax}(W_mh_n+b_m)
\end{equation}
where $h_n$ is the representation of $w_n$, the last token of $x$. Since $h_n$ captures context information of the entire sentence, we regard $p_n$ as the metaphorical probability of $x$ and use it to weight this training instance. 

\begin{figure*}[ht]
    \centering
    \includegraphics[width=0.8\textwidth]{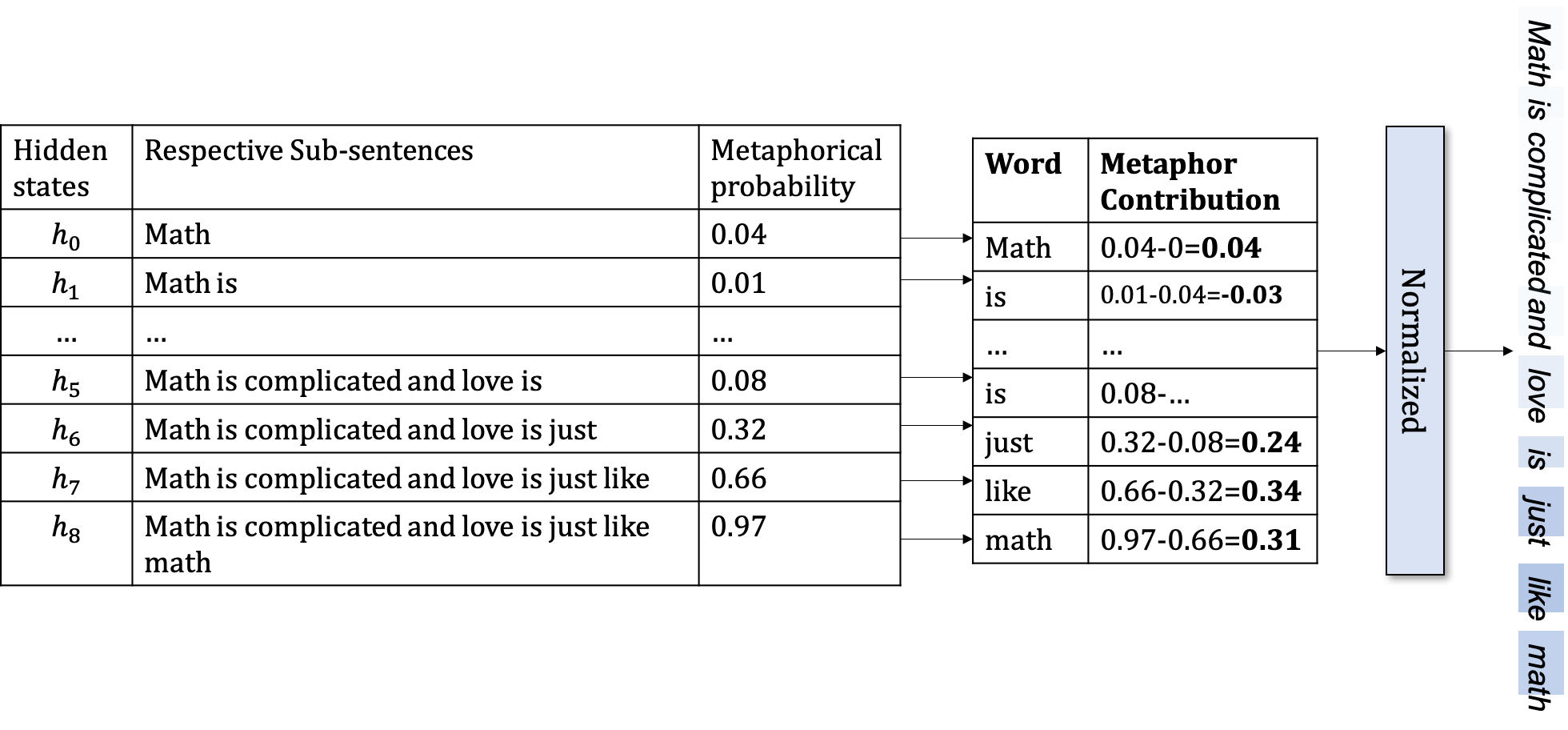}
    \caption{Metaphorical words identification by computing its contribution to metaphor probability.}
    \label{fig:phrases-ident}
\end{figure*}

\subsection{Metaphor Weighting} 
We find that in metaphorical sentences, only partial of the text is metaphor-related(i.e., the comparison between objects/concepts pair and comparator).
To verify, we analyse 200 metaphorical sentences randomly sampled from our dataset. We empirically find that for these metaphorical sentences, only 27\% of words are metaphor-related. This indicates that it is necessary to encourage our model to focus on metaphor-related parts of input metaphors; otherwise, it might lead to lower training efficiency and sub-optimal performance.

To tackle this issue, we propose a novel approach based on metaphor detection to identify metaphorical words within sentences. 
As shown in Figure \ref{fig:phrases-ident}, this is achieved by calculating the contribution to metaphor probability of each word to identify metaphor-related words. Formally, we use $I_i$ to indicate the importance of the $i$-th token in the metaphor modelling.
\begin{equation}
I_i=p_i-p_{i-1}
\end{equation}
where $p_{i-1}, p_i$ represents the metaphorical probability of subsentences $w_0, \cdots, w_{i-1}$ and $w_0, \cdots, w_i$. If $I_i$ is positive, it indicates the $i$-th token makes the sub-sentence $w_0, \cdots, w_i$ more metaphorical. On the contrary, if $I_i$ is zero or negative, it means this token is irrelevant to metaphorical expression. We weight each training step (i.e., given $w_0, \cdots, w_{i-1}$, predict $w_i$) with its corresponding $I'_i$, which is computed as follow: \begin{equation}
I'_i=\frac{\mathrm{exp}(\mathrm{max}(0, I_i))}{\sum_1^n \mathrm{exp}(\mathrm{max}(0, I_i))}
\end{equation}

\subsection{Training and Inference} \label{training and inference}

\paragraph{\textbf{Training}}
The training process of \textit{metaphor identification} is as follows. Given a labelled corpus $U=\{(x_i, y_i)\}_{i=1}^N$, where threetance consists of a sentence $x$ and a label $y$ indicating whether $x$ is a metaphorical sentence. We minimize the following loss function:
\begin{align}
L=&-\sum_{x\in U}\mathrm{log}P(\hat y|x)\\ \nonumber
=&-\sum_{x\in U}\mathrm{log}(\mathrm{softmax}(W_mh_n+b_m))
\end{align}
where $h_n$ is the representation of the last token of $x$.

The training procedure of \textit{metaphor modelling} is as follows. Given an unlabelled dataset $C=\{x_i\}_{i=1}^N$ where each instance is a sentence $x=(w_0, \cdots, w_n)$ with a list of words. Each instance is weighted by its metaphorical probability and each token $w_i$  is weighted by $I_i$, 
i.e., the metaphorical probability of word $w_i$. 
Formally, the loss function of our metaphor modelling is given as follows:
\begin{equation}
\small
L=-\sum_{x\in U\cup C}P(\hat y|x)\sum_i^n I'_i\cdot \mathrm{log}P(w_i|w_0,\cdots,w_{i-1})
\end{equation}

\paragraph{\textbf{Inference}} At the inference stage, we regard our model as a normal metaphorical language model to generate metaphors and do not perform metaphor identification. Given a target word $w_t$, i.e., the subject of the metaphor, we feed the target word concatenated with delimiter as input to our model. Our model produces the next word recurrently until the \textsc{EndofSentence} token is generated.

\section{Experiment}

\subsection{dataset}
To train our multitask framework, we construct two datasets: a labeled Chinese Metaphor Corpus (CMC) and a large-scale unlabeled Chinese Literature Corpus (CLC).

\noindent \textbf{CMC dataset} \quad
Existing Chinese metaphor corpus are neither too small (e.g., \citet{su2016-meta-interpretation} contains only 120 examples), or focus on a specific comparison, such as \citet{liu2018-simile-detect} which only consider metaphors with a specific comparator \zh{像} (like). We annotate Chinese Metaphor Corpus (CMC) to cover as many diverse metaphor patterns as possible.  CMC contains a total of 2703 metaphors and 3554 ordinary sentences, and is used to train our model to detect potential metaphors.

The 6.3k examples are sampled from 47,000 annotated sentences from Chinese literature corpus. We recruited three native Chinese annotators to perform the metaphor annotation task. Before annotation, we provide annotation guidelines to annotators and explain relevant definitions with metaphor examples. The annotation task is to judge whether a sentence is a metaphor. Each sentence is labelled by three annotators and the final label takes the majority of the three labelled results. The annotation agreement rate of CMC is 0.84 based on Krippendorff's alpha \cite{krippendorff2011-iaa} statistics. The statistics of CMC are shown in Table~\ref{tab:dataset}, along with some metaphor examples given in Table~\ref{tab:cmc-example}.

\vspace{2mm}
\noindent \textbf{CLC dataset} \quad In self-training, we need a large-scale corpus to enable the metaphor identification module detecting novel NMs. However, popular Chinese corpora, such as news, Wikipedia, web pages, are not suitable to be used as metaphor resources. Intuitively, literature text might be a promising resource of diverse metaphors. Therefore, we construct a Chinese literature corpus by collecting a large number of essays, novels, and fictions (see details in Appendix \ref{appendix:clc}). The statistics of CLC are shown in Table \ref{tab:dataset}.
\begin{table}[t]
    \centering
    \begin{tabular}{lcc}
    \hline
         & CMC & CLC \\ \hline
         \# Sentences & 6257 & 6.98M \\ 
         \# Metaphors & 2787 & - \\ 
         \# literal sentence & 3554 & - \\ 
         \#  tokens & 225K & 202M \\ 
         \# tokens per sentence & 35 & 29 \\ \hline
    \end{tabular}
    \caption{Statistics of CMC and CLC datasets}
    \label{tab:dataset}
\end{table}

\begin{table}[t]
    \centering
    \begin{CJK}{UTF8}{gbsn}
    \begin{tabular}{lp{4.5cm}}
\hline
    Type & Examples \\ \hline
    Metaphor & 瀑布注入水潭的一刹那,一朵朵白色的一浪一花腾空而起,像溅玉抛珠一般。\newline At the moment when the waterfall was poured into the pool, a white spray of flowers vacated, like a splash of jade beads. \\ \hline
    Not metaphor & 泛着银光的大海在身后铺展开来。\newline The silver-filled sea spread out behind him.
 \\ \hline
\end{tabular}
\end{CJK}
    \caption{Examples of metaphor and not metaphor in CMC.}
    \label{tab:cmc-example}
\end{table}


\subsection{Baselines}
Chinese metaphor generation is a novel task. We select three general generative models and an English simile generation method as baselines.

\noindent\textbf{RNN}: A LSTM based auto-regressive generative model, which consists of three LSTM layers.

\noindent\textbf{SeqGAN: } Sequence Generative adversarial network \cite{yu2017-seqgan} with a generator implemented by LSTM network and a discriminator implemented by CNN network. We train this model on CMC to produce Chinse metaphor.

\noindent\textbf{GPT2: } The Chinese GPT2 model is fine-tuned on the CMC dataset to produce Chinese metaphors as a baseline model.

\noindent\textbf{BART: } We construct parallel data \textit{(target word, metaphor)} from CMC and use the paired data to fine-tune a Chinese version BART model \cite{shao2021cpt} model.

\noindent \textbf{SCOPE: } \cite{chakrabarty2020-simile-gen} A SOTA method on English simile generation tasks, which fine-tunes BART model on a large-scale automatically created literal-simile parallel corpus.

\begin{table*}[t]
    \centering
    \begin{tabular}{lccccccc}
    \toprule
    & \multicolumn{4}{c}{\textbf{Automatic Eval}} & \multicolumn{3}{c}{\textbf{Human Eval}} \\
    \cmidrule(lr){2-5} \cmidrule(lr){6-8}
       \textbf{Methods} & \textbf{PPL} & \textbf{Dist-1}  & \textbf{Dist-2} & \textbf{Meta} & \textbf{Fluency} & \textbf{Consistency} & \textbf{Creativity} \\ \midrule
       RNN & 45.617 & .00758 & .1564 & \textbf{.955} & 1.60 (.58) & 2.25 (.42) & 2.65 (.31)  \\
       SeqGAN & 89.43 & .00336 & .0116 & \textbf{.998} & 3.33 (.51) & 3.80 (.46) & 1.67 (.34)  \\
       GPT2 & 57.88 & .00916 & .1154 & .981  & 4.00 (.62) & 3.10 (.39) & 2.60 (.31) \\
       BART & 48.58 & .00826 & .0971 & .978 & 4.35 (.54) & 3.05 (.37) & 2.30 (.32) \\
       SCOPE & 92.32 & .00517 & .0673 & .910 & 3.10 (.64) & 2.70 (.44) & 2.10 (.45) \\
        \midrule
       Our Method & \textbf{26.79} & \textbf{.01143} & \textbf{.1582} & .952  & \textbf{4.55} (.58) & \textbf{4.23} (.45) & \textbf{3.80} (.36) \\ 
       w/o Self-training & 62.54 & .00674 & .0906 & .982 & 3.85 (.54) & 3.87 (.42) & 2.76 (.38)  \\
       w/o Identification & 27.58 & .01050 & .1529 & .803 & 4.50 (.63) & 3.91 (.32) & 3.41 (.43) \\ \bottomrule
       
    \end{tabular}
    \caption{Results of automatic metrics and human evaluation. Boldface denotes the best results among our method and baselines. The inter-annotator agreement for human evaluation are shown in parenthesis.}
    \label{tab:results}
\end{table*}

\subsection{Experiments Setting}
We use a pre-trained Chinese GPT2 model\footnote{\url{https://huggingface.co/uer/gpt2-chinese-cluecorpussmall}} to avoid starting training from scratch. Our model is pre-trained on metaphor identification task with CMC for 3 epochs before jointly optimizing the two task-specific loss functions.  The implementation of SeqGAN\footnote{\url{https://github.com/LantaoYu/SeqGAN}} and the pre-trained Chinese BART model\footnote{\url{https://huggingface.co/fnlp/bart-base-chinese}} can be found in the footnote. Before training the RNN model and SeqGAN models on CMC, we first pre-train the RNN and the generator of SeqGAN on CLC for 50k steps. 
Hyper-parameters not specified are all followed by pytorch default settings.
Note that the SCOPE model is designed for English Simile generation and it takes a literal utterance as input. To compare SCOPE results with our method, we first translate the input target word into English using Google Translator, and then translate the generated outputs back to Chinese (details given in Appendix \ref{appendix:scope}). In the test stage, we randomly select and feed 200 target words from CMC to all generative models. During decoding, all beam sizes are set to 12, thus each model generated 12 sentence for each target. In total, 2400 sentences are obtained per model for testing.

\subsection{Metrics}
\noindent \textbf{Automatic Metrics} \quad We use perplexity (\textbf{PPL}) to evaluate the fluency of the generated text, which is calculated by an open source Chinese language model \cite{zhang2020cpm}. \textbf{Dist-1} and \textbf{Dist-2} \cite{li2016-dist} compute the distinct unigrams and bigrams ratio of generated text, which are used to measure model's ability to produce diversity outputs. To test the \textbf{metaphoricity (Meta)} of generated outputs, we train a RoBERTa-based Chinese metaphor classifier on CMC to compute the ratio of metaphorical utterances in the generated sentences. The accuracy of this classifier is 97.89\%, which is robust enough to perform evaluation. We show details of the classifier in Appendix \ref{appendix:meta}.

\noindent \textbf{Human Evaluation} \quad Due to the creative and delicate usage of metaphor, automatic metrics are not adequate to test the quality of generated outputs. We also perform human evaluation based on the following three criteria: 1) \textbf{Fluency} indicates how well the metaphor is formed; whether the expression is grammatical and fluent. 2) \textbf{Consistency} indicates whether the metaphor make sense; how well the subject of the metaphor related to the object. 3) \textbf{Creativity} scores how creative annotators think the metaphor is. Note that the Creativity judgment is based on annotators' real-life experience, rather than measuring whether the generated metaphor appears in the training dataset. Three annotators were instructed to rate the three criteria from 1 to 5, where 1 denotes worst and 5 be the best.

\begin{table*}[t]
    \centering
    \resizebox{\textwidth}{!}{
    \begin{tabular}{lp{6cm}p{5.5cm}cc}
    \hline
        \textbf{Methods} & \textbf{Text} (Chinese) & \textbf{Text} (Translated) & \textbf{Con.} & \textbf{Cre.}   \\ \hline
        GPT2 & \zh{秋天是美丽的，让人赏心悦目。} & Autumn is beautiful, and is delightful to the eye. & - & - \\
        & \zh{山是翠绿的，像一块无暇的宝石。} & Mountain is green just like an innocent jade. & 4.0 & 3.5 \\
        & \zh{爱情就像是蜂蜜，蜜蜂采蜂蜜。} & Love is like honey, bees collect honey. & 2.3 & 1.3 \\ \hline
        SCOPE & \zh{秋天象征春天，像一个月前。} & Autumn is a symbol of spring, like a month ago. & - & - \\
        & \zh{山是激情的，就像一个爱人。} & Mountain is a symbol of passion, like a lover. & 1.3 & 2.3 \\
        & \zh{爱情是浪漫的，像一盒巧克力。} & Love is a symbol of romance, like a box of chocolate. & 3.3 & 2.7 \\ \hline
        Our method & \zh{秋风是那么轻，从远处飘来，就像一条银色的绸带，拂着你，使你感到舒畅，心生向往}  & The autumn wind is so light, floating from a distance, like a silver ribbon, brushing against you, making you feel comfortable and yearning & 5.0 & 3.7 \\
         & \zh{山是那神的眼睛，像一盏盏璀璨明灯。} & Mountains are the eyes of the god, like bright lights. & 3.3 & 5.0 \\
         & \zh{恋爱像赌博，我就像赌场的赌徒，赌到手了就赢了，结果却输了。} & Love is like gambling, I was the gambler, if I win I will get you, but I lost. & 4.3 & 4.0 \\ \hline
    \end{tabular}
    }
    \caption{Example metaphors generated by our method and baselines. \textbf{Con.} and \textbf{Cre.} indicate the two human evaluation metrics \textbf{Consistency} and \textbf{Creativity} respectively. We do not assign Con. and Cre. score for non-metaphorical utterances. More examples of our method are shown in Appendix \ref{appendix:examples}.}
    \label{tab:cases}
\end{table*}

\section{Results}
\subsection{Automatic Evaluation}
Results of automatic metrics are shown in Table \ref{tab:results}. Our method significantly outperforms baselines in most automatic metrics. Our model obtains a lower PPL, which illustrates our model is better at producing fluency and grammatical text. Higher Dist-1 and Dist-2 scores show our method produces less repetitive unigrams and bigrams during generation, which is essential in creative language generation. The Meta (metaphor) score shows that our model produces more literal expressions than baselines, which might result from the self-training procedure, where non-metaphorical sentences are sometimes wrongly identified by the metaphor identification module, and hence introduces some noise in the metaphor learning process.

We implemented an ablation study to test the effectiveness of self-training and metaphor weighting. Experimental results show the effectiveness of the self-training mechanism in improving both generation fluency and diversity. It can also be observed that removing self-training from our model affects four automatic metrics by a large margin. The metaphor weighting mechanism mainly helps improve the metaphoricity of the generated metaphors and thus improves the Meta score.

\subsection{Human Evaluation}
We select 180 sentences in total for human evaluation, 
where the results are shown in Table \ref{tab:results}. The Table also shows the inter-annotator agreement of human annotation via Krippendorff's alpha. We can see that our method beats five baseline models on all three human-centric metrics. The most significant improvement lies in Consistency and Creativity, which shows our method not only can generate creative comparisons, but also can provide a consistent context for each nominal metaphor, which is essential for readability and explainability. Human evaluation also demonstrates the effectiveness of self-training. Self-training enhances generation quality in both fluency and creativity aspects. Metaphor weighting mechanism shows less effectiveness in human evaluation as metaphor weighting mainly aims to improve the metaphoricity of the generated output and human metrics do not measure this aspect. We provide a visualisation for metaphor weighting in Appendix \ref{app:visualize}.

\subsection{Case Study}
We show some generated examples of GPT2, SCOPE, and our model  in Table \ref{tab:cases}, where the corresponding Consistency and Creativity scores are also provided. Specifically, models generate metaphors by taking different target word as the input. We see that although all three models are able to produce metaphorical outputs, the quality of the generated results differs among systems. \textbf{First}, in some cases, baseline model seem fail to generate metaphorical outputs. For example, when feeding ``\textit{autumn}'' as input to GPT2 and the SCOPE model, both fail to produce a metaphor but a literal description. \textbf{Second}, the comparisons given by our model are more creative than baselines, where GPT2 and SCOPE tend to generate with some common metaphor patterns. For example, GPT2 compares mountain with ``\textbf{jade}'', which is a very common metaphor in Chinese. \textbf{Finally}, we find our method generates metaphors in a relatively more complicated structure and speaks in a more poetic way. For example, our method does not employ a single word in constructing comparison; instead, it tend to generate 
detailed phrases such as ``\textit{love is a gambling and I was the gambler}'', ``\textit{autumn wind is like a ribbon brush against you}''. These detailed components paint a more vivid picture, and thus improve the overall readability of the metaphors. The corresponding human-rated Consistency and Creativity scores also support this observation.

\section{Conclusion}
In this paper, we introduce a novel neural metaphor generator and construct the first Chinese metaphor corpus. The proposed model can effectively learn potential metaphors in unlabelled corpus by self-training and emphasise metaphor-related words in the metaphor modelling process with metaphorical word identification. Experimental results show that our method is able to generate metaphors of good readability and quality with limited labelled training data. In future work, we plan to generate metaphor with richer linguistic constructs and extend our approach to metaphor generation of other languages.

\bibliography{anthology,custom}
\bibliographystyle{acl_natbib}

\appendix

\section{Chinese Literature Corpus (CLC)}
\label{appendix:clc}
CLC consists of three main categories of Chinese literature: Children's Literature (Children), Chinese Literature (Chinese), Translated Literature (Translated). Statistics of each category are shown in Table \ref{tab:clc}.

\begin{table}[h]
\centering
\begin{tabular}{lccc}
\hline
    \textbf{Category} & \textbf{\#Books} & \textbf{\#Tokens} & \textbf{\#Sentences} \\\hline
    Children & 195 & 17M & 0.58M \\ 
    Chinese & 336 & 64M & 2.2M \\
    Translated & 854 & 121M & 4.2M\\ \hline
\end{tabular}
\centering
\caption{Summary of CLC.}
\label{tab:clc}
\end{table}

\section{SCOPE Model}
\label{appendix:scope}

SCOPE model takes a literal expression as input and produces a simile correspondingly. For example, given ``the city is beautiful'', SCOPE model will transfer the literal expression into a simile: ``The city is like a painting''. 

In our experiments, to compare SCOPE with our method, we first 1) feed a \textsc{tenor} to COMET \cite{bosselut2019comet} model, to get properties of the \textsc{tenor}. For example, given a query ``<Autumn, SymbolOf>'', COMET predicts a list of properties for Autumn: ``Passion, gold'' etc. We then 2) construct literal expressions using the \textsc{tenor} and its properties. For example, ``Autumn is a symbol of passion'' is obtained. 3) The literal expression is fed to SCOPE model and a simile is produced. For example, ''Autumn is like a lover'' is produced by SCOPE model. 4) At last, the simile are concatenate with its literal expression to form a complete NM with context: ''Autumn is a symbol of passion, like a lover''.

\section{Meta Metric}
\label{appendix:meta}

The CMC corpus is splited into training set (80\%) and test set (20\%) for training the classifier. We simply add a linear layer plus a binary softmax layer on the RoBERTa model as the NM classifier. The accuracy of the classifier tested on test set of CMC is 97.89\%.
\begin{table*}[!t]
    \centering
    \begin{tabular}{p{0.45\textwidth}p{0.45\textwidth}}
    \hline
        Text (Chinese) & Text (Translated) \\ \hline
        \zh{爱像一缕金光，即使在黑夜也能照亮你的心灵。} &  Love is like a ray of golden light, which can illuminate your heart even at night. \\
        \zh{爱像一盏明亮的夜灯，让迷途的航船找到港湾；} & Love is like a bright night light, let the lost ship find the harbor. \\
        \zh{时间像利剑一样无情的锋刃，一旦出鞘，瞬间就割断你人生的纽带。} & Time is a ruthless blade like a sharp sword. Once it comes out of the scabbard, it will cut off the bond of your life in an instant. \\
        \zh{秋天像个美人的画笔调侃着大地：世界上再没有比这更美的了。} & Autumn teases the earth like a beautiful brush: there is nothing more beautiful in the world. \\ 
        \zh{爱心像一片照射在冬日的陽光，使饥寒交迫的人感到人间的温暖.} & Love is like a piece of sunshine in winter, which makes hungry and cold people feel the warmth of the world \\\hline
    \end{tabular}
    \caption{More generation examples of MetaGen.}
    \label{tab:more_examples}
\end{table*}
\section{More Examples}
\label{appendix:examples}

Table \ref{tab:more_examples} shows generations produced by our method given different \textsc{tenors}.

\section{Visualization of Metaphor Weighting}
\label{app:visualize}
The visualization of metaphor weighting mechanism is shown in Table \ref{tab:visualization}.

\begin{table*}[ht]
    \centering
    \small
    \begin{tabular}{c|lp{5.5cm}}
    \hline
       \textbf{Meta Score}  & \textbf{Examples} & \textbf{Examples (Translated)}  \\ \hline
         0.984 & \begin{CJK*}{UTF8}{gbsn}{\setlength{\fboxsep}{0pt}
\colorbox{red!59.550}{\strut 煤}%
\colorbox{red!28.660}{\strut 油}%
\colorbox{red!74.587}{\strut 灯}%
\colorbox{red!47.020}{\strut 光}%
\colorbox{red!36.157}{\strut ，}%
\colorbox{red!83.412}{\strut 仿}%
\colorbox{red!94.653}{\strut 佛}%
\colorbox{red!43.665}{\strut 是}%
\colorbox{red!13.592}{\strut 大}%
\colorbox{red!100.000}{\strut 海}%
\colorbox{red!49.288}{\strut 湾}%
\colorbox{red!47.849}{\strut 里}%
\colorbox{red!42.025}{\strut 的}%
\colorbox{red!29.013}{\strut 渔}%
\colorbox{red!89.928}{\strut 灯}%
\colorbox{red!58.723}{\strut 野}%
\colorbox{red!50.762}{\strut 火}%
\colorbox{red!0.000}{\strut 。}%
}\end{CJK*} & The kerosene lamp is like a fishing lamp wildfire in the bay. \\
     0.893 & \begin{CJK*}{UTF8}{gbsn}{\setlength{\fboxsep}{0pt}
\colorbox{red!0.000}{\strut 大}%
\colorbox{red!44.781}{\strut 公}%
\colorbox{red!84.431}{\strut 豺}%
\colorbox{red!49.038}{\strut 皮}%
\colorbox{red!38.611}{\strut 毛}%
\colorbox{red!37.645}{\strut 亮}%
\colorbox{red!35.829}{\strut 得}%
\colorbox{red!100.000}{\strut 像}%
\colorbox{red!11.812}{\strut 天}%
\colorbox{red!59.813}{\strut 边}%
\colorbox{red!56.546}{\strut 的}%
\colorbox{red!53.489}{\strut 云}%
\colorbox{red!41.279}{\strut 霞}%
。
}\end{CJK*} & The fur of the cormorant is like the clouds in the sky. \\
    0.887 & \begin{CJK*}{UTF8}{gbsn}{\setlength{\fboxsep}{0pt}
\colorbox{red!60.996}{\strut 这}%
\colorbox{red!46.025}{\strut 声}%
\colorbox{red!44.040}{\strut 音}%
\colorbox{red!41.291}{\strut 就}%
\colorbox{red!100.000}{\strut 像}%
\colorbox{red!41.952}{\strut 无}%
\colorbox{red!0.000}{\strut 形}%
\colorbox{red!67.874}{\strut 的}%
\colorbox{red!59.615}{\strut 闪}%
\colorbox{red!97.563}{\strut 电}%
\colorbox{red!79.075}{\strut 一}%
\colorbox{red!76.783}{\strut 般}%
。
}\end{CJK*} & This sound is like invisible lightning. \\
    0.736 & \begin{CJK*}{UTF8}{gbsn}{\setlength{\fboxsep}{0pt}
\colorbox{red!19.168}{\strut 它}%
\colorbox{red!56.960}{\strut 们}%
\colorbox{red!55.583}{\strut 很}%
\colorbox{red!100.000}{\strut 像}%
\colorbox{red!54.401}{\strut 一}%
\colorbox{red!30.856}{\strut 对}%
\colorbox{red!78.037}{\strut 孪}%
\colorbox{red!81.114}{\strut 生}%
\colorbox{red!77.665}{\strut 兄}%
\colorbox{red!96.091}{\strut 弟}%
。
}\end{CJK*} & They are much like twin brothers.\\
    0.213 & \begin{CJK*}{UTF8}{gbsn}{\setlength{\fboxsep}{0pt}
\colorbox{red!20.935}{\strut 漂}%
\colorbox{red!23.367}{\strut 泊}%
\colorbox{red!12.365}{\strut ，}%
\colorbox{red!23.421}{\strut 会}%
\colorbox{red!34.239}{\strut 让}%
\colorbox{red!0.000}{\strut 他}%
\colorbox{red!34.000}{\strut 见}%
\colorbox{red!54.658}{\strut 识}%
\colorbox{red!43.199}{\strut 到}%
\colorbox{red!27.787}{\strut 他}%
\colorbox{red!32.993}{\strut 没}%
\colorbox{red!22.015}{\strut 有}%
\colorbox{red!23.964}{\strut 见}%
\colorbox{red!32.327}{\strut 到}%
\colorbox{red!27.976}{\strut 过}%
\colorbox{red!30.738}{\strut 的}%
\colorbox{red!21.110}{\strut 东}%
\colorbox{red!38.839}{\strut 西}%
。
}\end{CJK*} & Wandering will let him see things he has never seen before. \\ \hline
    \end{tabular}
    \caption{Sentences in Chinese literature corpus with its metaphor probability and the visualization of weights in metaphor weighting mechanism are presented for each token.}
    \label{tab:visualization}
\end{table*}

\end{document}